\documentclass[runningheads]{llncs}
\usepackage{graphicx}
\usepackage[utf8]{inputenc}
\usepackage[OT4]{fontenc}
\usepackage{booktabs}
\usepackage{rotating}
\usepackage{url}
\usepackage{float}
\usepackage{hyperref}
\usepackage{multirow}

\begin{document}
\title{Keyword Extraction from Short Texts with~a~Text-To-Text Transfer Transformer}
\titlerunning{Keyword Generation with a Text-To-Text Transfer Transformer (T5)}

\author{Piotr Pęzik\inst{1,2}\orcidID{0000-0003-0019-5840} \and \\
Agnieszka Mikołajczyk\inst{2}\orcidID{0000-0002-8003-6243} \and \\
Adam Wawrzyński\inst{2}\orcidID{0000-0002-1698-2390} \and \\
Bartłomiej Nitoń\inst{3}\orcidID{0000-0003-3306-7650}  \and \\
Maciej Ogrodniczuk\inst{3}\orcidID{0000-0002-3467-9424}}

\authorrunning{P. Pęzik, A. Mikołajczyk, A. Wawrzyński, B. Nitoń, M. Ogrodniczuk}
%
\institute{University of Łódź, Faculty of Philology 
\and
VoiceLab, NLP Lab
\and
Institute of Computer Science, Polish Academy of Sciences
}
\maketitle              
\begin{abstract}
The paper explores the relevance of the Text-To-Text Transfer Transformer language model (T5) for Polish (plT5) to the task of intrinsic and extrinsic keyword extraction from short text passages. The evaluation is carried out on the new Polish Open Science Metadata Corpus (POSMAC), which is released with this paper: a collection of  216,214 abstracts of scientific publications compiled in the CURLICAT project. We compare the results obtained by four different methods, i.e. plT5kw, extremeText, TermoPL, KeyBERT and conclude that the plT5kw model yields particularly promising results for both frequent and sparsely represented keywords. Furthermore, a plT5kw keyword generation model trained on the POSMAC also seems to produce highly useful results in cross-domain text labelling scenarios. We discuss the performance of the model on  news stories and phone-based dialog transcripts which represent text genres and domains extrinsic to the dataset of scientific abstracts. Finally, we also attempt to characterize the challenges of evaluating a text-to-text model on both intrinsic and extrinsic keyword extraction. 
\end{abstract}


\section{Keyword Extraction and Generation}

The main NLP problem discussed in this paper can be described as keyword extraction or generation from short text passages. More specifically, given a span of text such as a concatenated title and abstract of a research paper, the task is to generate a small set of words or multiword phrases (usually nominal phrases) which succinctly describe its content. Approaches to this problem can be purely \textit{extractive} or partly  \textit{abstractive}. In the former case, keywords are extracted and possibly normalized more or less directly from the text of the sample. Abstractive methods can assign labels that may not have occurred in the original text sample or a restricted vocabulary of known keywords. 

Despite the long tradition of keyword extraction as a distinct NLP task, "no single approach (...) effectively extracts keywords from different data sources" \cite{firoozeh2020keyword}\footnote{This paper also offers an up-to-date review of keyword extraction methods.}. Extraction of keywords from longer texts can be approached similarly to term extraction and partly facilitated by considering word frequency distributions and identifying significantly frequent phrases as potential keywords. This paper focuses mostly on showcasing the applications of the T5 generative language model \cite{DBLP:journals/corr/abs-1910-10683} and comparing its performance to text classification (extremeText / fastText) \cite{wydmuch2018noregret}\cite{joulin2016bag}  and statistical terminology extraction (C / NC-values) \cite{frantzi_automatic_2000} as baseline methods. Although the complementarity of statistical and transformer-based approaches to keyword extraction has been explored before \cite{10.1007/978-3-030-79150-6_50}, we are not aware of any published  assessment of text-to-text generative models on this task.

From the point of view of model evaluation, it should be noted that the manual assignment of keywords to scientific articles, which are used as groundtruth annotations in our analysis, is far from deterministic and quite different from text classification or labelling based on a closed-set taxonomy. Authors usually draw a small set of terms from a largely uncontrolled vocabulary. Such descriptors may be terminological items used in the text of their paper, but they can also be more abstract or at least hyperonymic descriptors of its content. Synonyms, hyperonyms  and abbreviated forms contribute to the apparent sparsity of the vocabulary, which over time tends to grow in a large collection of abstracts at a sublinear rate. This in turn has implications for building and evaluating automatic keyword extraction solutions. Firstly, the distribution of keywords as distinct class labels in many datasets is rather sparse, which means that the recall of rare keywords is unlikely to be high, at least in any supervised text classification scenario. Secondly, the evaluation of automatically assigned keywords is problematic as the `ground truth` assignments are neither consistent nor exhaustive. The latter problem could be systematically addressed by measuring inter-rater agreement in datasets which are explicitly developed for NLP purposes. However, the corpus of scientific abstracts used in this study has been adapted from metadata sources which were not globally curated and checked for consistency. 

Despite these methodological limitations, we believe that our evaluation of keyword generation approaches provides fresh insights into the transferability of a T5 model to loosely related topical domains and text genres. As shown in the last section of this paper, a model tuned on a high-quality corpus of scientific abstract extracts surprisingly accurate keywords from news stories and even spoken dialogue transcripts.
Thus, the novelty of this paper consists in the fact that a) we test the relevance of text-to-text transfer transformers to the task of keyword generation and b) we evaluate and release a non-obvious dataset which shows significant potential of transferability to extrinsic domains and languages.

\section{The Polish Open Science Metadata Corpus}
The source dataset used in this study was developed in CURLICAT\footnote{\url{https://curlicat.eu/}}, an international project aimed at delivering rich metadata monolingual corpora in seven EU languages, including Polish, and representing different topical domains and text genres. The Polish subset of CURLICAT (released for the first time with this paper) named Polish Open Science Metadata Corpus (POSMAC)\footnote{\url{http://clip.ipipan.waw.pl/POSMAC}} --- contains a new source of valuable corpus data acquired from the Library of Science (LoS)\footnote{\url{https://bibliotekanauki.pl/}}, a platform providing open access to full texts of articles published in over 900 Polish scientific journals and full texts of selected scientific books together with extensive bibliographic metadata. More than 70~\% of the metadata records included in the resulting corpus contain keywords describing the content of the indexed articles.  Since authors of the respective articles typically enter such keywords themselves, their selection is relatively uncontrolled. After lowercasing and ASCII-folding (i.e., removing Polish diacritics due to their inconsistent use), we found a total of 256 139 distinct keywords used in the corpus, with only 69 266 (ca. 27\%) used more than once and 10 074 keywords assigned to 10 or more articles. Syntactically, the vast majority of keywords are lemmatized noun phrases whose length typically ranges from 1 to 3 words (mean=2.39, sd=1.22). A single article record is tagged with an average of 4.76 keywords (median=4).

\begin{table}[]
\caption{Top 10 scientific domains represented in the POSMAC.}
\renewcommand{\arraystretch}{1.15}
\label{tab:docsDomains}
\begin{tabular}{lrp{0.3cm}r}
\toprule
\multirow{2}{*}{\textbf{Domains}} & \multirow{2}{*}{\textbf{Documents}} & & \multicolumn{1}{c}{\textbf{With}} \\
& & & \multicolumn{1}{c}{\textbf{keywords}} \\
\midrule
Engineering and technical sciences                       & 58 974 & & 57 165 \\
Social sciences                                          & 58 166 & & 41 799 \\
Agricultural sciences                                    & 29 811 & & 15 492 \\
Humanities                                               & 22 755 & & 11 497 \\
Exact and natural sciences                               & 13 579 & &  9 185 \\
Humanities, Social sciences                              & 12 809 & &  7 063 \\
Medical and health sciences                              &  6 030 & &  3 913 \\
Medical and health sciences, Social sciences             &    828 & &    571 \\
Humanities, Medical and health sciences, Social sciences &    601 & &    455 \\
Engineering and technical sciences, Humanities           &    312 & &    312 \\
\bottomrule
\end{tabular}
\end{table}

\section{Approaches}

\subsection{T5, plT5 and plT5kw}
T5 stands for the Text-To-Text Transfer Transformer model proposed by \cite{DBLP:journals/corr/abs-1910-10683}. In terms of its architecture, the model is based on the original encoder-decoder transformer implementation \cite{vaswani2017attention}. Unlike popular transformer-based language models used in classification tasks, T5 frames all NLP problems as text-to-text operations, where both the input and output are text strings. Although this approach may only seem natural for selected NLP problems such as question answering, translation or summarization, it has been demonstrated to apply to other tasks such as classification or regression tasks. In this study, the input to a T5 model is a concatenated title and abstract of a scientific paper and the text string output is a comma-separated list of lemmatized single- or multiword `keywords'\footnote{We use the traditional term \textit{keyword} to refer to potentially multiword phrases found in the \textit{Keywords} section of a scientific abstract.}. In the case of morphologically rich languages such as Polish, such lemmatization may additionally involve number, case and gender agreement operations on the resulting multiword keywords. This requirement is particularly important for out-of-vocabulary (OOV) keywords which need to be lemmatized and formatted on demand. 

For the extraction of Polish keywords, we used the plT5-base model \cite{plt5}\footnote{\url{https://huggingface.co/allegro/plt5-large}} trained on six reference corpora of Polish. More specifically, we train the model to predict comma-separated keywords from article abstracts concatenated with titles. We used an Adam optimizer with 100 warm-up steps, linearly increasing the learning rate from zero to a target of 3e-5. Additionally, we used a multiplicative scheduler that lowered the LR by 0.7 every epoch. We trained the model for ten epochs with a batch size of 8. The maximum input length was set to 512 tokens and the maximum target length was 128. We refer to the resulting keyword extraction model as \textbf{plT5kw}. 

We experimented with \texttt{no\_repeat\_ngram\_size} and \texttt{num\_beams} parameters on \textit{dev} subset of datasets to find out the best configuration. During evaluation on the test subset, we set \texttt{no\_repeat\_ngram\_size} to 3 and \texttt{num\_beams} to 4.

\begin{figure}[h]
    \centering
    \includegraphics[width=\textwidth]{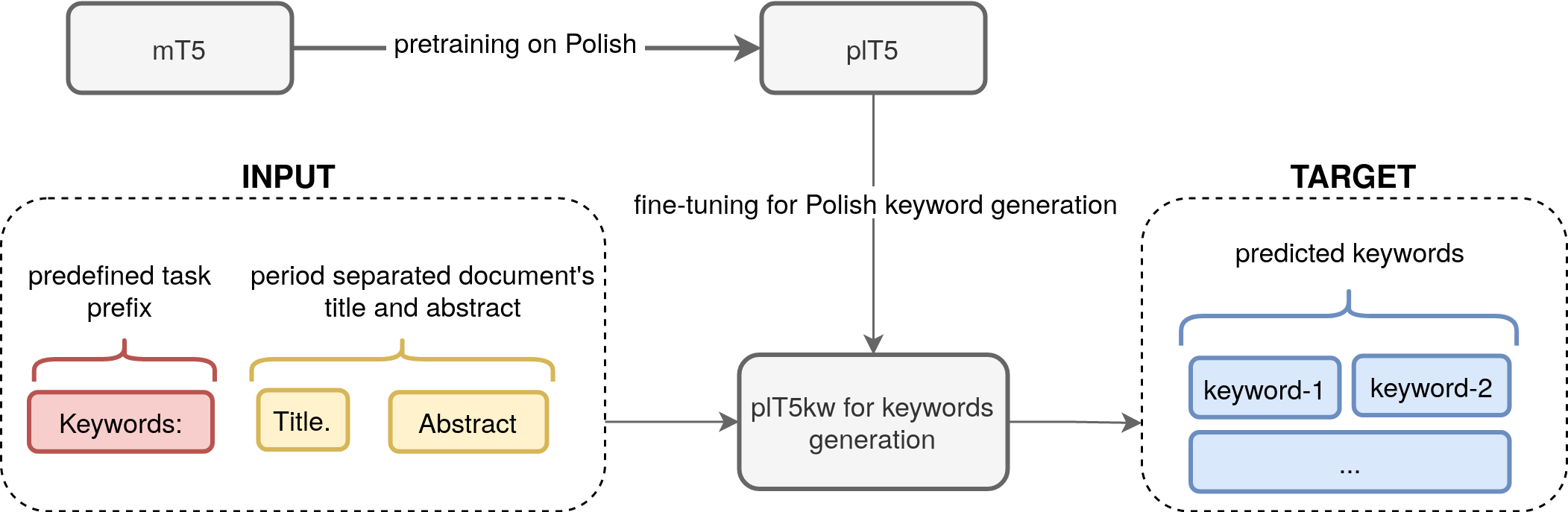}
    \caption{Training procedure for our Text-To-Text Transfer Transformer model for keywords generation.}
    \label{fig:keyt5}
\end{figure}

\subsection{FastText and extremeText}

FastText is a popular text classification library which uses vector representations of (sub)words as input to a relatively simple neural network \cite{joulin2016bag}. Despite the obvious differences between supervised text classification and unsupervised keyword extraction, the assignment of keywords attested in a representative collection of tagged texts can be treated as a text labelling task. In the comparison described in this paper we used an extension of FastText called    
extremeText \cite{wydmuch2018noregret}, which uses a Probabilistic Labels Tree loss function (PLT) to optimize the assignment of labels from very large taxonomies, such as the set of over 200 000 distinct keywords found in POSMAC. Using the PLT loss function, we trained a  keyword classifier with 300 dimensions in the hidden layer for 50 epochs to obtain the results reported below. 

\subsection{TermoPL}

TermoPL \cite{mar:myk:rych:lrec16} is a statistical terminology extraction tool designed to identify recurrent words and multiword combinations in domain corpora of Polish. It identifies, lemmatizes and scores recurrent noun phrases as potential terminological items using a ranking function proposed by \cite{frantzi_automatic_2000}. We include this approach to measure the upper bound of purely extractive keyword identification. In other words, we estimate the maximum recall of simply extracting and lemmatizing all noun phrases contained in any given abstract.

\subsection{KeyBERT}

KeyBERT \cite{grootendorst2020keybert} is a keyword extraction library utilizing BERT representations. For each document it creates a representation vector using a transformer-based language model. Next, word representations for each n-gram found in a given text are compared with the document vector using cosine vector distance scores. The most similar phrases are selected as those that represent the document content. Additionally, two methods are used to increase the diversity of the generated phrases: Maximum Sum Similarity and Maximal Marginal Relevance. We used KeyBERT with the \textit{distiluse-base-multilingual-cased-v1} model from the Sentence Transformers library \cite{reimers-2020-multilingual-sentence-bert}. The filtering method used in the experiments was Maximal Margin Relevance, and the diversity factor was set to 0.7 with an n-gram range of (1,2). The other parameters were used with their default values.




\section{Intrinsic Evaluation}

To evaluate the above-mentioned set of complementary keyword generation approaches intrinsically (i.e. on the original POSMAC dataset), abstracts annotated with keywords were split into a training and test set with a ratio of 70/30\%. To ensure consistent distribution of labels in the training and test set, we used an implementation of the iterative stratification algorithm\footnote{See \url{https://vict0rs.ch/2018/05/24/sample-multilabel-dataset/}.} for multilabel data, originally proposed by \cite{sechidis2011stratification}. The relevance and coverage of keyword assignments are evaluated in terms of micro- and macro- precision and recall values, as well as their harmonic means (F\textsubscript{1}-scores) averaged over the documents in the test set. These scores are measured at several ranks (k=1, 3, 5 and more) for each approach in two different scenarios: a) using the full set of keywords assigned in the training and test set and b) training and/or evaluating only on keywords which occur at least 10 times in the stratified dataset.

It should be noted that in addition to evaluating the four main approaches described in this paper, we also assessed several baseline keyword extraction approaches, including  FirstPhrases and TopicRank \cite{bougouin-etal-2013-topicrank}, PositionRank \cite{florescu-caragea-2017-positionrank}, MultipartiteRank \cite{https://doi.org/10.48550/arxiv.1803.08721}, TextRank \cite{mihalcea-tarau-2004-textrank}, KPMiner \cite{el-beltagy-rafea-2010-kp} and TfIdf with some adjustments aimed at boosting their performance (such as lemmatizing input text). The results obtained for all of those methods were less than 0.025 F$_1$ on all ranks, which is why we excluded them from the detailed comparisons included below. 

 Table \ref{tab:int-eval-full-set} compares the performance of extremeText and plT5kw on the task of extracting keywords from the full set of more than 200 000 items found in POSMAC. The highest average F\textsubscript{1} (harmonic mean of precision and recall) is obtained for both approaches at rank 5, although plT5kw clearly outperforms extremeText on this task both in terms of precision and recall at all the corresponding ranks. We include rank 10 for the extremeText classifier but not for plT5kw, because the former can be requested to produce a ranked list of keywords used in the training set of any length while the latter is implicitly trained to produce text strings with up to 4 commas on average. The results for KeyBERT for Polish keyword extraction are very poor both in terms of recall and precision, neither of which increases over 0.03 at any rank measured (see the qualitative explanations below). As signalled above, TermoPL is meant to be used on longer texts as it ranks terms by a score which is partly derived from their frequency in a sufficiently large corpus of texts. Nevertheless, looking at its recall when no rank limit is applied, it is interesting to observe that more than 33 percent of all keywords in our corpus are actually some form of nominal phrases found in the text of the abstracts.    
\begin{table}[H]
\setlength{\tabcolsep}{8pt}
\renewcommand{\arraystretch}{1.15}
\centering
\caption{Results of evaluation on the full set of POSMAC keywords. The upper part of the table presents the results for all keywords present in the dataset, while the lower part refers to experiments conducted with the rejection of keywords occurring less than 10 times.}
\label{tab:int-eval-full-set}
\begin{tabular}{lrrrrrrr}
\toprule
\multirow{2}{*}{\textbf{Method}} & \multirow{2}{*}{\textbf{Rank}} & \multicolumn{3}{c}{\textbf{Micro}} & \multicolumn{3}{c}{\textbf{Macro}} \\ 
 & & \multicolumn{1}{c}{\textbf{P}} & \multicolumn{1}{c}{\textbf{R}} & \multicolumn{1}{c}{\textbf{F$_1$}}  & \multicolumn{1}{c}{\textbf{P}} & \multicolumn{1}{c}{\textbf{R}} & \multicolumn{1}{c}{\textbf{F$_1$}}   \\ 
\midrule
\multirow{4}{*}{extremeText}                       & 1\hspace{5pt}  & 0.175           & 0.038        & 0.063    & 0.007           & 0.004        & 0.005      \\
                                                   & 3\hspace{5pt}  & 0.117           & 0.077        & 0.093    & 0.011           & 0.011        & 0.011       \\
                                                   & 5\hspace{5pt}  & 0.090           & 0.099        & 0.094    & 0.013           & 0.016        & 0.015      \\
                                                   & 10\hspace{5pt} & 0.060           & 0.131        & 0.082    & 0.015           & 0.025        & 0.019      \\ 
\midrule
\multirow{3}{*}{plT5kw}                              & 1\hspace{5pt}  & \textbf{0.345}               & 0.076            & 0.124        & 0.054               & 0.047            & 0.050 \\
                                                   & 3\hspace{5pt}  & 0.328               & 0.212            & 0.257        & 0.133               & 0.127            & 0.129 \\
                                                   & 5\hspace{5pt}  & 0.318               &  \textbf{0.237}            &  \textbf{0.271}        & 0.143               & 0.140            & 0.141 \\ 
\midrule
\multirow{3}{*}{KeyBERT}                       & 1\hspace{5pt}  & 0.030           & 0.007        & 0.011    & 0.004           & 0.003        & 0.003      \\
                                                   & 3\hspace{5pt}  & 0.015           & 0.010        & 0.012    & 0.006           & 0.004        & 0.005       \\
                                                   & 5\hspace{5pt}  & 0.011           & 0.012        & 0.011    & 0.006           & 0.005        & 0.005    \\ 
\midrule
\multirow{4}{*}{TermoPL}                       & 1\hspace{5pt}  & 0.118           & 0.026        & 0.043    & 0.004           & 0.003        & 0.003      \\
                                                  & 3\hspace{5pt}  & 0.070           & 0.046        & 0.056    & 0.006           & 0.005        & 0.006       \\
                                                  & 5\hspace{5pt}  & 0.051           & 0.056        & 0.053    & 0.007           & 0.007        & 0.007      \\
                                                  & all\hspace{3pt} & 0.025           & 0.339        & 0.047    & 0.017           & 0.030        & 0.022      \\ 
\toprule
\multirow{4}{*}{extremeText}                       & 1\hspace{5pt}  & 0.210           & 0.077        & 0.112    & 0.037           & 0.017        & 0.023     \\
                                                   & 3\hspace{5pt}  & 0.139           & 0.152        & 0.145    & 0.045           & 0.042        & 0.043      \\
                                                   & 5\hspace{5pt}  & 0.107           & 0.196        & 0.139    & 0.049           & 0.063        & 0.055   \\
                                                   & 10\hspace{5pt} & 0.072           & 0.262        & 0.112    & 0.041           & 0.098        & 0.058      \\ \midrule
\multirow{3}{*}{plT5kw}                              & 1\hspace{5pt}  &  \textbf{0.377}               & 0.138            & 0.202       & 0.119               & 0.071            & 0.089 \\
                                                   & 3\hspace{5pt}  & 0.361               & 0.301            & 0.328        & 0.185               & 0.147            & 0.164 \\
                                                   & 5\hspace{5pt}  & 0.357               &  \textbf{0.316}            &  \textbf{0.335}        & 0.188               & 0.153            & 0.169 \\ \midrule
\multirow{3}{*}{KeyBERT}                       & 1\hspace{5pt}  & 0.018           & 0.007        & 0.010    & 0.003           & 0.001        & 0.001      \\
                                                   & 3\hspace{5pt}  & 0.009           & 0.010        & 0.009    & 0.004           & 0.001        & 0.002       \\
                                                   & 5\hspace{5pt}  & 0.007           & 0.012        & 0.009    & 0.004           & 0.001        & 0.002    \\ \midrule
\multirow{4}{*}{TermoPL}                       & 1\hspace{5pt}  & 0.076           & 0.028        & 0.041    & 0.002           & 0.001        & 0.001      \\
                                                  & 3\hspace{5pt}  & 0.046           & 0.051        & 0.048    & 0.003           & 0.001        & 0.002        \\
                                                  & 5\hspace{5pt}  & 0.033           & 0.061        & 0.043    & 0.003           & 0.001        & 0.002       \\
                                                  & all\hspace{3pt} & 0.021           & 0.457        & 0.040    & 0.004           & 0.008        & 0.005 \\
\bottomrule
\end{tabular}
\end{table}

The lower part of Table \ref{tab:int-eval-full-set} shows the results of evaluating the four approaches on a set of 10,083 distinct keywords which were assigned to at least 10 different abstracts in the stratified dataset. As expected, the results for extremeText are slightly better in this run as they are not affected by very rare or unknown keywords in the test set. However, the tuned plT5kw is again significantly better in this scenario.

Table \ref{tab:int-eval-invt} reveals an interesting property of plT5kw. Its precision increases up to 0.425 at rank 1 when the predictions are limited to keywords found in the training set. The improvement in precision comes at the expense of recall. which drops by nearly 10 percentage points. This observation also means that unlike extremeText or any text classification model plT5kw is capable of assigning relevant keywords which were not seen in the training set. The transferability of plT5kw is further discussed in the next section of this paper.

\begin{table}[H]
\centering
\renewcommand{\arraystretch}{1.15}
\setlength{\tabcolsep}{8pt}
\caption{Evaluation of plT5kw on the set of keywords found in the training set.}
\label{tab:int-eval-invt}
\begin{tabular}{lcrrrrrr}
\toprule
\multirow{2}{*}{\textbf{Method}} & \multirow{2}{*}{\textbf{Rank}} & \multicolumn{3}{c}{\textbf{Micro}} & \multicolumn{3}{c}{\textbf{Macro}} \\ 
 & & \multicolumn{1}{c}{\textbf{P}} & \multicolumn{1}{c}{\textbf{R}} & \multicolumn{1}{c}{\textbf{F$_1$}}  & \multicolumn{1}{c}{\textbf{P}} & \multicolumn{1}{c}{\textbf{R}} & \multicolumn{1}{c}{\textbf{F$_1$}}   \\ 
\midrule
\multirow{3}{*}{plT5kw}                              & 1  & 0.425               & 0.093            & 0.153       & 0.086               & 0.074            & 0.080 \\
                                                   & 3  & 0.415               & 0.212            & 0.281        & 0.165               & 0.158            & 0.161 \\
                                                   & 5  & 0.412               & 0.227            & 0.293        & 0.172               & 0.167            & 0.169 \\ \bottomrule
\end{tabular}
\end{table}

\section{Transfer to Other Domains}

Although the results of intrinsic evaluation of keyword extraction from scientific abstracts reported above may seem moderately useful, the plT5kw model trained on a rather narrowly defined source domain seems to produce surprisingly precise (although incomplete) keywords for samples of other topical domains and text genres. In this section, we  explore the relevance of the plT5kw model trained on POSMAC to the domain of news stories and transcripts of conversational speech. We also compare the type of keywords produced by the four extraction approaches discussed above in more qualitative terms. 

\subsection{News Stories}

Table \ref{tab:news_results} shows a set of shorthand English translations of headlines of recent news stories published in Polish web-based media outlets. The full text of each story is linked to the shorthand headline. The next four columns of the table show samples of keywords generated for the full text of each article by the four respective extraction methods described in this paper. 

The overall quality of the extracted keywords can be considered from a number of perspectives. The \textbf{importance} of an extracted keyword (also known as \textbf{keyness}) refers to its potential to express the most important aspects of a text passage. Although all important keywords are also \textbf{relevant} (related to the content of the text sample), not all relevant keywords are equally important and usually some limit on the \textbf{complete} set of relevant keywords is required. The \textbf{abstraction} aspect of keywords pertains to the degree to which they can describe the content without necessarily relying on the verbatim word combinations used in a given text passage. The \textbf{transferability} of a keyword generation method refers to its ability to produce good quality keywords for texts from  domains which are different from those of the originally labelled datasets. Finally, the \textbf{formatting} quality of a keyword refers to the correct lemmatization, true-casing or abbreviation of the extracted keywords.

\begin{sidewaystable}
\centering
\caption{Comparison of keywords generated for 5 news stories.}
\label{tab:news_results}
\begin{tabular}[t]{@{}llllll@{}}
\toprule
\textbf{\#} & \textbf{Short title}  & \textbf{plT5kw}                       & \textbf{KeyBERT} & \textbf{TermoPL} & \textbf{ExtremeText}                                          \\ \midrule
1           & \begin{tabular}[c]{p{0.18\textwidth}} \href{https://www.polsatnews.pl/wiadomosc/2021-12-12/w-niedziele-spotkanie-premiera-morawieckiego-z-kanclerzem-niemiec-olafem-scholzem/}{Polish PM meets new} \\
\href{https://www.polsatnews.pl/wiadomosc/2021-12-12/w-niedziele-spotkanie-premiera-morawieckiego-z-kanclerzem-niemiec-olafem-scholzem/}{German Chancellor}
\end{tabular} & \begin{tabular}[c]{p{0.17\textwidth}}Niemcy,\\ Polska,\\ Unia Europejska,\\ Nord Stream 2,\\ bezpieczeństwo,\\ kryzys migracyjny,\\ polityka\end{tabular} & \begin{tabular}[c]{p{0.2\textwidth}}premiera morawieckiego, \\ migracyjne energetyczne, \\ powiedzieli czego,  \\ będzie bardzo, \\ serii spotkań\end{tabular}         & \begin{tabular}[c]{p{0.2\textwidth}}nowy kanclerz Niemiec,\\ instrument szantażu, \\ kwestia migracyjna, \\ kanclerz Niemiec, \\ Olaf Scholza, ... \\  (237 in total)\end{tabular} & \begin{tabular}[c]{p{0.14\textwidth}}rosja, \\  polska, \\  unia\_europejska \end{tabular} \\
\midrule
2           & \begin{tabular}[c]{p{0.18\textwidth}}\href{https://wiadomosci.onet.pl/swiat/wybuch-gazociagu-na-sycylii-sa-ofiary/nex6pn7}{Gas pipe explosion} \\\href{https://wiadomosci.onet.pl/swiat/wybuch-gazociagu-na-sycylii-sa-ofiary/nex6pn7}{on Sicily}\end{tabular}  & \begin{tabular}[c]{p{0.17\textwidth}}gaz,\\ wybuch,\\ Włochy\end{tabular}                                                                                 & \begin{tabular}[c]{p{0.2\textwidth}}wybuchu gazociągu, \\   czterech osób, \\   miejscowości ravanusa, \\   jak poinformowała,  \\  pomoże przeszukiwaniu\end{tabular} & \begin{tabular}[c]{p{0.2\textwidth}}dziesiątek osób, \\   miejsce wybuchu, \\   wybuch gazociągu, \\   włoska wyspa, ... \\  (71 in total)\end{tabular}                            & \begin{tabular}[c]{p{0.14\textwidth}}paleontologia, \\  skamienialosci, \\  fauna\_kopalna \end{tabular} \\
\midrule
3           & \begin{tabular}[c]{p{0.18\textwidth}}\href{https://www.rmf24.pl/fakty/polska/news-atak-na-kieleckim-rynku-policja-szuka-napastnika,nId,5702938#crp_state=1}{Person wounded}\\\href{https://www.rmf24.pl/fakty/polska/news-atak-na-kieleckim-rynku-policja-szuka-napastnika,nId,5702938#crp_state=1}{in Kielce knife attack}\end{tabular}                                                                & \begin{tabular}[c]{p{0.17\textwidth}}Kielce,\\ policja,\\ rana,\\ rynek\end{tabular}                                                                      & \begin{tabular}[c]{p{0.2\textwidth}}policji kielcach, \\   rynku 31, \\   wyjaśniają okoliczności, \\   narzędziem przez, \\   letni mężczyzna\end{tabular}            & \begin{tabular}[c]{p{0.2\textwidth}}kielecki rynek, \\   poszukiwanie napastnika,  \\  ostre narzędzie, ... \\  (44 in total)\end{tabular}                                         & \begin{tabular}[c]{p{0.14\textwidth}}bezpieczenstwo, \\  kontrola, \\  policja\end{tabular} \\
\midrule
4           & \begin{tabular}[c]{p{0.18\textwidth}}\href{https://www.polsatnews.pl/wiadomosc/2021-12-12/wielka-brytania-pierwsze-przypadki-hospitalizacji-osob-zakazonych-wariantem-omikron/}{Omicron coronavirus} \\\href{https://www.polsatnews.pl/wiadomosc/2021-12-12/wielka-brytania-pierwsze-przypadki-hospitalizacji-osob-zakazonych-wariantem-omikron/}{variant in UK}\end{tabular}        & \begin{tabular}[c]{p{0.17\textwidth}}Covid,\\ Omikron,\\ Wielka Brytania,\\ koronawirus,\\ hospitalizacja\end{tabular}                                    & \begin{tabular}[c]{p{0.2\textwidth}}przypadki hospitalizacji, \\ powiedzmy 50, \\ minister edukacji, \\ brytania pierwsze, \\ liczba potwierdzonych\end{tabular}       & \begin{tabular}[c]{p{0.2\textwidth}}nowy wariant, \\   wariant Omikron, \\   wielka Brytania, ... \\  (126 in total)\end{tabular}                                                  & \begin{tabular}[c]{p{0.14\textwidth}}migracja, \\  bezpieczenstwo, \\  transport\_kolejowy\end{tabular} \\
\midrule
5           & \begin{tabular}[c]{p{0.18\textwidth}}\href{https://www.rmf24.pl/fakty/polska/news-wandale-pomalowali-granitowy-glaz-nad-morskim-okiem,nId,5703025#crp_state=1
}{Tatry National Park} \\\href{https://www.rmf24.pl/fakty/polska/news-wandale-pomalowali-granitowy-glaz-nad-morskim-okiem,nId,5703025#crp_state=1
}{landmark vandalized}\end{tabular} & \begin{tabular}[c]{p{0.17\textwidth}}Tatry,\\ wandalizm,\\ szlak turystyczny\end{tabular}                                                                 & \begin{tabular}[c]{p{0.2\textwidth}}pomalowali granitowy, \\ narodowy film, \\ tatrzański park, \\ morskiego oka,  \\ sprawę gdzie\end{tabular}                        & \begin{tabular}[c]{p{0.2\textwidth}}morskie oko, \\   tatrzański park narodowy, \\   granitowy głaz, ... \\  (88 in total)\end{tabular}                                            & \begin{tabular}[c]{p{0.14\textwidth}}historia, \\  ochrona, \\  bezpieczenstwo\end{tabular} \\ \bottomrule
\end{tabular}
\end{sidewaystable}

The transferability of the keywords produced by \textbf{extremeText}, a closed-label set classifier is clearly limited. The predictions are far from complete and they are only remotely relevant to the texts from the news domains in cases where some overlap exists between the original domain of scientific abstracts and a given news story. The labels are static, i.e. they are not dynamically reformatted or adjusted to the text samples.

The results produced by the \textbf{KeyBERT} model are probably the least convincing in this comparison. The model shows a clear preference for longer n-grams, which may not be syntactically complete nominal phrases or any regular phrases for that matter. The results are not lemmatized or properly cased, but their relevance is at least relatively easy to judge as they can be traced back to the exact span of text from which they were extracted.

As a terminology extraction solution \textbf{TermoPL} produces  lemmatized although not always correctly cased noun phrases. Some of the results are clearly complementary to the keywords produced by plT5kw, but the choice of the most important items remains an issue as this solution requires a larger body of text to score its suggestions. The solution is domain-independent, but this also means that it does not transfer any knowledge about the desired keyword format or level of abstraction from other domains.

The keywords produced by \textbf{plT5kw} are mostly relevant and well abstracted although occasionally also too generic. For example the meeting of the Polish PM with the German Chancellor is tagged as \textit{Poland} and \textit{Germany} and the story about a gas explosion in Sicily gets the rather generic tag of \textit{Italy}.  The model does not seem to have a bias toward single- or multiword expressions. It produces correctly lemmatized, syntactically agreed and well-cased phrases. It seems to have transferred the skill of identifying, formatting and sometimes abstracting comma-separated nominal phrases without over-fitting excessively to the topics represented in the source domain. Needless to say, the recall of the model is far from perfect, but its precision is reliably stable.

\subsection{Customer Support Dialogues}

The promising results obtained on a sample corpus of news stories have led us to test plT5kw on the completely different domain of phone conversation transcripts which were sampled from the DiaBiz corpus.\footnote{DiaBiz is a corpus developed in the CLARIN-Biz project. It contains some 4,000 phone-based customer support calls covering a range of topics and business processes.} The following excerpt from the DiaBiz corpus is a translation of a dialogue between a support agent informing a client about an outstanding electricity bill which has resulted in an energy supply disconnection. Our plT5kw model trained on scientific abstracts labels the following passage with two noun phrases: \textit{loan} and \textit{financial advisory} even though there is only a handful of abstracts with these keywords in the POSMAC:

\begin{figure}
\textbf{Customer}: I can't imagine how I could live without electricity. I need to use my fridge and washing machine. I guess I have no... I don't even know where I could borrow some money. Is there any way I could pay my debt in some kind of installments. How should I go about it? \\[5pt]
\textbf{Agent}: It's alright. I understand and... I'm really sorry about your situation. Before we continue, however, we need to sort out a few formal issues. Can I ask you again to state your name and email address? \\[5pt]
\end{figure}

Table \ref{tab:diabiz-domains-results} shows frequencies of keywords assigned to a set of 50 DiaBiz transcripts representing scenarios from 6 different customer support domains. The recurrent keywords seem to accurately (if not exhaustively) summarize the underlying conversations. \textit{Logistyka} (\textit{logistics}) is the only potentially irrelevant keyword in this subset which may have resulted from some over-fitting of the model on the original domain of scientific abstracts.

\begin{table}[H]
\renewcommand{\arraystretch}{1.15}
\setlength{\tabcolsep}{8pt}
\centering
\caption{Frequent keywords generated by plT5kw for phone dialogues in different customer support domains and their English translations.}
\label{tab:diabiz-domains-results}
\begin{tabular}{lll}
\toprule
\textbf{Domain} & \textbf{Keywords PL} & \textbf{Keywords EN} \\ 
\midrule
medical     & gastroskopia~(4)         & gastroscopy (4)       \\
            & diagnostyka medyczna~(3) & medical diagnosis (3) \\
            & numer PESEL~(2)          & National Identification Number~(2) \\ 
            & diagnostyka~(2)          & diagnosis (2)         \\
            & lekarz POZ~(1)        & GP~(1)                 \\[3pt]
\midrule

tourism     & hotel~(9)                & hotel~(9)           \\     
            & turystyka~(6)            & tourism~(6)             \\     
            & Afryka~(2)               & Africa~(2)            \\     
            & Turcja~(2)               & Turkey~(2)            \\     
            & atrakcje~(2)              & attractions~(2)        \\[3pt]
\midrule

insurance   & naprawa~(6)              & repair~(6)            \\     
            & uszkodzenie~(5)          & damage~(5)            \\     
            & ekspres do kawy~(3)      & coffee machine~(3)    \\     
            & reklamacja~(3)           & complaints~(3)        \\     
            & ubezpieczenie~(2)         & insurance~(2)          \\
\bottomrule
\end{tabular}
\end{table}

\begin{table}[H]
\renewcommand{\arraystretch}{1.15}
\setlength{\tabcolsep}{8pt}
\centering
\begin{tabular}{lll}
\toprule
\textbf{Domain} & \textbf{Keywords PL} & \textbf{Keywords EN} \\ 
\midrule
            
banking     & identyfikacja~(6)        & identification~(6)    \\      
            & bankowość internetowa~(4)& online banking~(4)    \\       
            & logistyka~(3)            & logistics~(3)         \\      
            & Bank Narodowy S.A.~(3)   & National Bank S.A.~(3)\\      
            & logowanie~(2)             & login~(2) \\[3pt]      
\midrule
energy      & licznik~(6)              & meter~(6)                \\                                                
            & zerwanie plomby~(4)      & seal break~(4)           \\                                                
            & plomba na liczniku~(8)   & seal on meter~(8)        \\                                                
            & PESEL~(2)                & National Identification Number~(2) \\
            & weryfikacja tożsamości~(1)& identity verification~(1) \\
\bottomrule
\end{tabular}
\end{table}

\section{Summary and Future Work}

Our evaluation of a keyword extraction solution based on a text-to-text transformer shows that the fine-tuned model called plT5kw outperforms the other approaches when tested on the original dataset of scientific abstracts. Furthermore, a preliminary analysis of keywords assigned to text from very different domains (news stories and speech transcripts) shows that the proposed solution is capable of generating relevant, properly formatted and well-abstracted keywords on extrinsic text samples. One of the limitations of this study stems from the fact that manual keyword annotations are intrinsically biased against high recall evaluations as authors are artificially restricted to assign a limited number of terms to each text. Therefore, a more systematic quantitative evaluation on extrinsic domains would require manually annotated datasets verified for inter-rater agreement. We envisage further challenges which need to be addressed in future research on this problem. For example, it seems reasonable to assume that open-set keyword extraction could benefit from distributional vector-based techniques of normalizing semantically equivalent keywords. Also, there are potential benefits of zero- or few-shot fine-tuning of text-to-text keyword extraction models to the target domain, which need to be considered more systematically. Finally, the results obtained in this study may vary for  different languages, which requires further evaluation, possibly on multilingual variants of the T5 model used.

\section*{Acknowledgements}

The work reported here was supported by 1) the European Commission in the CEF Telecom Programme (Action No: 2019-EU-IA-0034, Grant Agreement No: 	
INEA/CEF/ICT/A2019/1926831) and the Polish Ministry of Science and Higher Education: research project 5103/CEF/2020/2, funds for 2020–2022) and 2) the National Centre for Research and Development, research grant POIR.01.01.01-00-1237/19.

\bibliographystyle{splncs04}
\bibliography{aciids} 

\begin{thebibliography}{10}
\providecommand{\url}[1]{\texttt{#1}}
\providecommand{\urlprefix}{URL }
\providecommand{\doi}[1]{https://doi.org/#1}

\bibitem{https://doi.org/10.48550/arxiv.1803.08721}
Boudin, F.: Unsupervised {K}eyphrase {E}xtraction with {M}ultipartite {G}raphs.
  In: Proceedings of the 2018 Conference of the North {A}merican Chapter of the
  Association for Computational Linguistics: Human Language Technologies,
  Volume 2 (Short Papers). pp. 667--672. Association for Computational
  Linguistics, New Orleans, Louisiana (2018),
  \url{https://aclanthology.org/N18-2105}

\bibitem{bougouin-etal-2013-topicrank}
Bougouin, A., Boudin, F., Daille, B.: {T}opic{R}ank: Graph-based topic ranking
  for keyphrase extraction. In: Proceedings of the Sixth International Joint
  Conference on Natural Language Processing. pp. 543--551. Asian Federation of
  Natural Language Processing, Nagoya, Japan (2013),
  \url{https://aclanthology.org/I13-1062}

\bibitem{plt5}
Chrabrowa, A., Dragan, {\L}., Grzegorczyk, K., Kajtoch, D., Koszowski, M.,
  Mroczkowski, R., Rybak, P.: Evaluation of {T}ransfer {L}earning for {P}olish
  with a {T}ext-to-{T}ext {M}odel. In: Proceedings of the 13th Conference on
  Language Resources and Evaluation ({LREC} 2022). pp. 4374--4394. European
  Language Resources Association, Marseille, France (2022),
  \url{http://www.lrec-conf.org/proceedings/lrec2022/pdf/2022.lrec-1.466.pdf}

\bibitem{el-beltagy-rafea-2010-kp}
El-Beltagy, S.R., Rafea, A.: {KP}-miner: Participation in {S}em{E}val-2. In:
  Proceedings of the 5th International Workshop on Semantic Evaluation. pp.
  190--193. Association for Computational Linguistics, Uppsala, Sweden (Jul
  2010), \url{https://aclanthology.org/S10-1041}

\bibitem{firoozeh2020keyword}
Firoozeh, N., Nazarenko, A., Alizon, F., Daille, B.: Keyword {E}xtraction:
  {I}ssues and {M}ethods. Natural Language Engineering  \textbf{26}(3),
  259--291 (2020)

\bibitem{florescu-caragea-2017-positionrank}
Florescu, C., Caragea, C.: {P}osition{R}ank: An unsupervised approach to
  keyphrase extraction from scholarly documents. In: Proceedings of the 55th
  Annual Meeting of the Association for Computational Linguistics (Volume 1:
  Long Papers). pp. 1105--1115. Association for Computational Linguistics,
  Vancouver, Canada (Jul 2017), \url{https://aclanthology.org/P17-1102}

\bibitem{frantzi_automatic_2000}
Frantzi, K., Ananiadou, S., Mima, H.: Automatic {R}ecognition of {M}ulti-word
  {T}erms: {T}he {C}-value/{NC}-value {M}ethod. International Journal on
  Digital Libraries  \textbf{3}(2),  115--130 (2000).
  \doi{10.1007/s007999900023}

\bibitem{10.1007/978-3-030-79150-6_50}
Giarelis, N., Kanakaris, N., Karacapilidis, N.: A {C}omparative {A}ssessment of
  {S}tate-of-the-art {M}ethods for {M}ultilingual {U}nsupervised {K}eyphrase
  {E}xtraction. In: Maglogiannis, I., Macintyre, J., Iliadis, L. (eds.)
  Artificial Intelligence Applications and Innovations. pp. 635--645. Springer
  International Publishing, Cham (2021)

\bibitem{grootendorst2020keybert}
Grootendorst, M.: Key{BERT}: {M}inimal {K}eyword {E}xtraction with {BERT}
  (2020). \doi{10.5281/zenodo.4461265}

\bibitem{joulin2016bag}
Joulin, A., Grave, E., Bojanowski, P., Mikolov, T.: Bag of {T}ricks for
  {E}fficient {T}ext {C}lassification. In: Proceedings of the 15th Conference
  of the {E}uropean Chapter of the Association for Computational Linguistics:
  Volume 2, Short Papers. pp. 427--431. Association for Computational
  Linguistics, Valencia, Spain (2017), \url{https://aclanthology.org/E17-2068}

\bibitem{mar:myk:rych:lrec16}
Marciniak, M., Mykowiecka, A., Rychlik, P.: {TermoPL} --- a {F}lexible {T}ool
  for {T}erminology {E}xtraction. In: Calzolari, N., Choukri, K., Declerck, T.,
  Grobelnik, M., Maegaard, B., Mariani, J., Moreno, A., Odijk, J., Piperidis,
  S. (eds.) Proceedings of the Tenth International {C}onference on {L}anguage
  {R}esources and {E}valuation ({LREC}~2016). pp. 2278--2284. European Language
  Resources Association (2016),
  \url{http://www.lrec-conf.org/proceedings/lrec2016/pdf/296_Paper.pdf}

\bibitem{mihalcea-tarau-2004-textrank}
Mihalcea, R., Tarau, P.: {T}ext{R}ank: Bringing order into text. In:
  Proceedings of the 2004 Conference on Empirical Methods in Natural Language
  Processing. pp. 404--411. Association for Computational Linguistics,
  Barcelona, Spain (2004), \url{https://aclanthology.org/W04-3252}

\bibitem{DBLP:journals/corr/abs-1910-10683}
Raffel, C., Shazeer, N., Roberts, A., Lee, K., Narang, S., Matena, M., Zhou,
  Y., Li, W., Liu, P.J.: Exploring the limits of transfer learning with a
  unified text-to-text transformer. Journal of Machine Learning Research
  \textbf{21}(140),  1--67 (2020), \url{http://jmlr.org/papers/v21/20-074.html}

\bibitem{reimers-2020-multilingual-sentence-bert}
Reimers, N., Gurevych, I.: Making {M}onolingual {S}entence {E}mbeddings
  {M}ultilingual using {K}nowledge {D}istillation. In: Proceedings of the 2020
  Conference on Empirical Methods in Natural Language Processing (EMNLP). pp.
  4512--4525. Association for Computational Linguistics (2020),
  \url{https://aclanthology.org/2020.emnlp-main.365/}

\bibitem{sechidis2011stratification}
Sechidis, K., Tsoumakas, G., Vlahavas, I.: On the {S}tratification of
  {M}ulti-label {D}ata. In: Gunopulos, D., Hofmann, T., Malerba, D.,
  Vazirgiannis, M. (eds.) Machine Learning and Knowledge Discovery in Databases
  ({ECML PKDD 2011}). pp. 145--158. Lecture Notes in Computer Science
  vol.~6913, Springer Berlin Heidelberg (2011).
  \doi{https://doi.org/10.1007/978-3-642-23808-6\_10}

\bibitem{vaswani2017attention}
Vaswani, A., Shazeer, N., Parmar, N., Uszkoreit, J., Jones, L., Gomez, A.N.,
  Kaiser, {\L}., Polosukhin, I.: Attention is {A}ll you {N}eed. In: Guyon, I.,
  von Luxburg, U., Bengio, S., Wallach, H.M., Fergus, R., Vishwanathan, S.V.N.,
  Garnett, R. (eds.) Advances in Neural Information Processing Systems 30:
  Proceedings of the Annual Conference on Neural Information Processing Systems
  (NeurIPS 2017). pp. 5998--6008 (2017),
  \url{https://proceedings.neurips.cc/paper/2017/hash/3f5ee243547dee91fbd053c1c4a845aa-Abstract.html}

\bibitem{wydmuch2018noregret}
Wydmuch, M., Jasinska, K., Kuznetsov, M., Busa-Fekete, R., Dembczy\'{n}ski, K.:
  A~{N}o-regret {G}eneralization of {H}ierarchical {S}oftmax to {E}xtreme
  {M}ulti-label {C}lassification. In: Proceedings of the 32nd International
  Conference on Neural Information Processing Systems (NeurIPS 2018). pp.
  6358--6368. Curran Associates Inc. (2018),
  \url{https://proceedings.neurips.cc/paper/2018/hash/8b8388180314a337c9aa3c5aa8e2f37a-Abstract.html}

\end{thebibliography}

\end{document}